\documentclass[10pt,twocolumn,letterpaper]{article}
\usepackage{iccv}
\usepackage{times}
\usepackage{epsfig}
\usepackage{graphicx}
\usepackage{amsmath}
\usepackage{amssymb}
\usepackage{algorithm}
\usepackage{algorithmic}
\usepackage[tight]{subfigure}
\usepackage{booktabs} 
\usepackage{amsthm}
\usepackage{color}
\usepackage{amsfonts}
\usepackage{array}
\usepackage{multirow}
\usepackage[pagebackref=true,breaklinks=true,letterpaper=true,colorlinks,bookmarks=false]{hyperref}

\theoremstyle{remark}

\theoremstyle{plain}
\newtheorem{theorem}{Theorem}

\newcolumntype{C}[1]{>{\centering\let\newline\\\arraybackslash\hspace{0pt}}m{#1}}

\iccvfinalcopy 

\def\iccvPaperID{2963} 

\ificcvfinal\fi
\begin{document}

\title{Generative Adversarial Minority Oversampling}

\author{Sankha Subhra Mullick\\
Indian Statistical Institute\\
Kolkata, India\\
{\tt\small sankha\_r@isical.ac.in}
\and 
Shounak Datta\\
Duke University\\
Durham, NC, USA\\
{\tt\small shounak.jaduniv@gmail.com}
\and
Swagatam Das\\
Indian Statistical Institute\\
Kolkata, India\\
{\tt\small swagatam.das@isical.ac.in}
}

\maketitle

\begin{abstract}
Class imbalance is a long-standing problem relevant to a number of real-world applications of deep learning. Oversampling techniques, which are effective for handling class imbalance in classical learning systems, can not be directly applied to end-to-end deep learning systems. We propose a three-player adversarial game between a convex generator, a multi-class classifier network, and a real/fake discriminator to perform oversampling in deep learning systems. The convex generator generates new samples from the minority classes as convex combinations of existing instances, aiming to fool both the discriminator as well as the classifier into misclassifying the generated samples. Consequently, the artificial samples are generated at critical locations near the peripheries of the classes. This, in turn, adjusts the classifier induced boundaries in a way which is more likely to reduce misclassification from the minority classes. Extensive experiments on multiple class imbalanced image datasets establish the efficacy of our proposal. 
\end{abstract}

\vspace{-0.3in}

\section{Introduction}\label{sec:intro}
The problem of class imbalance occurs when all the classes present in a dataset do not have equal number of representative training instances \cite{He2009learningimb,das2018handling}. Most of the existing learning algorithms produce inductive bias favoring the majority class in presence of class imbalance in the training set, resulting in poor performance on the minority class(es). This is a problem which routinely plagues many real-world applications such as fraud detection, dense object detection \cite{lin2017cost}, medical diagnosis, etc. For example, in a medical diagnosis application, information about unfit patients is scarce compared to that of fit individuals. Hence, traditional classifiers may misclassify some unfit patients as being fit, having catastrophic implications \cite{mazurowski2008training}.

\begin{figure}[!t]
\vskip 0.2in
\begin{center}
\subfigure[\label{fig:exampleMlp}]{\includegraphics[width=0.47\linewidth]{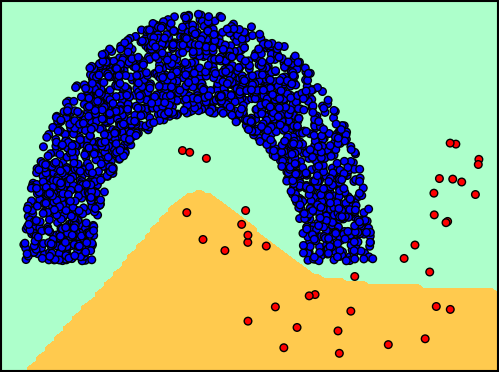}}
\hskip 0.03in
\subfigure[\label{fig:exampleCGan}]{\includegraphics[width=0.47\linewidth]{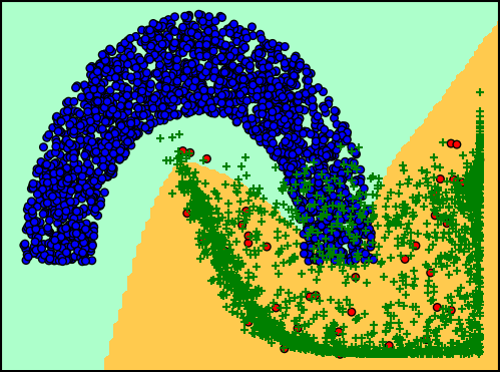}}
\subfigure[\label{fig:exampleGamoNoDis}]{\includegraphics[width=0.47\linewidth]{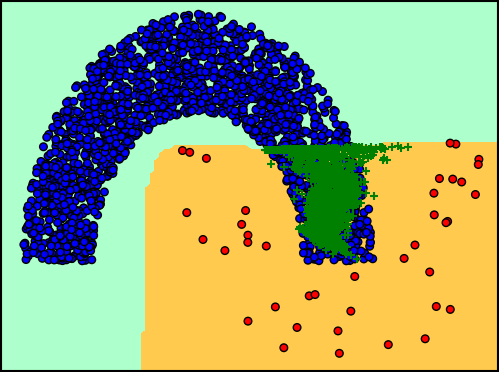}}
\hskip 0.03in
\subfigure[\label{fig:exampleGamo}]{\includegraphics[width=0.47\linewidth]{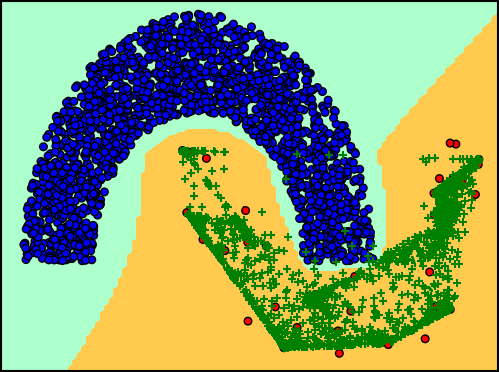}}
\caption{Illustration using a `toy' dataset: (a) Imbalanced classification with an unaided classifier network $M$ results in misclassification of the minority class instances (red dots). (b) Artificial minority points (green `+') generated using conditional GAN help to improve the result on the minority class but bleed into the majority class (blue dots), affecting the performance on the latter. (c) New points are generated by training a convex generator $G$ alternatingly with $M$. This is a two player adversarial game where $G$ attempts to generate samples which are hard for $M$ to correctly classify. This results in ideal performance on the minority class, but at the cost of misclassifying the majority class as $G$ does not adhere to the distribution of the minority class. (d) Ideal performance on both classes is achieved by further incorporating an additional discriminator $D$ to induce fidelity to the minority class distribution and to limit bleeding into majority class territory.}
\label{fig:example}
\end{center}
\vskip -0.4in
\end{figure}

Over the years, the machine learning community has devised many methods for tackling class imbalance \cite{krawczyk2016learning, branco2016}. However, only a few of these techniques have been extended to deep learning even though class imbalance is fairly persistent is such networks, severely affecting both the feature extraction as well as the classification process \cite{xie2015holistically, huang2016learning, xie2017holistically, buda2018systematic, khan2018cost}. The existing solutions \cite{huang2016learning, Chung2016c, wang2016d, lin2017cost, rota2017cost} for handling class imbalance in deep neural networks mostly focus on cost tuning to assign suitably higher costs to minority instances. Another interesting class of approaches \cite{yan2015, dong2018imbalanced} focuses on constructing balanced subsamples of the dataset. Wang et al. \cite{wang2017meta} proposed a novel meta-learning scheme for imbalanced classification. It is interesting to note that oversampling techniques like SMOTE \cite{chawla2002smote} have not received much attention in the context of deep learning, despite being very effective for classical systems \cite{fernandez2018smote}. This is because deep feature extraction and classification are performed in an end-to-end fashion, making it hard to incorporate oversampling which is typically done subsequent to feature extraction. An attempt to bridge this gap was made by Ando and Huang \cite{shin2017dos} in their proposed deep oversampling framework (DOS). However, DOS uniformly oversamples the entire minority class and is not capable of concentrating the artificial instances in difficult regions. Additionally, the performance of DOS depends on the choice of the class-wise neighborhood sizes, which must be determined by costly parameter tuning. 

Generative adversarial networks (GANs) are a powerful subclass of generative models that have been successfully applied to image generation. This is due to their capability to learn a mapping between a low-dimensional latent space and a complex distribution of interest, such as natural images \cite{goodfellow2014generative, mirza2014conditional, radford2015unsupervised, odena2017conditional}. The approach is based on an adversarial game between a generator that tries to generate samples which are similar to real samples and a discriminator that tries to discriminate between real training samples and generated samples. The success of GANs as generative models has led Douzas and Bacao \cite{douzas2018effective} to investigate the utility of using GANs to oversample the minority class(es). However, attempting to oversample the minority class(es) using GANs can lead to boundary distortion \cite{santurkar2018classification}, resulting in a worse performance on the majority class (as illustrated in Figure \ref{fig:exampleCGan}). Moreover, the generated points are likely to lie near the mode(s) of the minority class(es) \cite{srivastava2017veegan}, while new points around the class boundaries are required for learning reliable discriminative (classification) models \cite{hui2005bSmote, He2008adasyn}.

Hence, in this article, we propose (in Section \ref{sec:method}) a novel end-to-end feature-extraction-classification framework called Generative Adversarial Minority Oversampling (GAMO) which employs adversarial oversampling of the minority class(es) to mitigate the effects of class imbalance. The contributions made in this article differ from the existing literature in the following ways:
\vspace{-0.1in}
\begin{enumerate}
    \item Unlike existing deep oversampling schemes \cite{shin2017dos,douzas2018effective}, GAMO is characterized by a three-player adversarial game among a convex generator $G$, a classifier network $M$, and a discriminator $D$.
    \vspace{-0.07in}
    \item Our approach is fundamentally different from existing adversarial classification schemes (where the generator works in harmony with the classifier to fool the discriminator) \cite{salimans2016improved,kumar2017semi,springenberg2015unsupervised,odena2017conditional}, in that our convex generator $G$ attempts to fool both $M$ and $D$.
    \vspace{-0.07in}
    \item Unlike the generator employed in GAN \cite{goodfellow2014generative}, we constrain $G$ to conjure points within the convex hull of the class of interest. Additionally, the discriminator $D$ further ensures that $G$ adheres to the class distribution for non-convex classes. Consequently, the adversarial contention with $M$ pushes the conditional distribution(s) learned by $G$ towards the periphery of the respective class(es), thus helping compensate for class imbalance effectively.
    \vspace{-0.07in}
    \item In contrast to methods like \cite{chawla2002smote,douzas2018effective}, $G$ can oversample different localities of the data distribution to different extents based on the gradients obtained from $M$.
    \vspace{-0.07in}
    \item For applications requiring a balanced training set of images, we also propose a technique called GAMO2pix (Section \ref{sec:imgGen}) that can generate realistic images from the synthetic instances generated by GAMO in the distributed representation space.
    \vspace{-0.05in}
\end{enumerate}
We undertake an ablation study as well as evaluate the performance of our method compared to the state-of-the-art in Section \ref{sec:exp}, and make concluding remarks in Section \ref{sec:concl}.

\vspace{-0.05in}

\section{Related Works}
The success of SMOTE \cite{chawla2002smote, chawla2003smoteboost} has inspired several improvements. For example, \cite{hui2005bSmote, chumphol2009sls} attempt to selectively oversample minority class points lying close to the class boundaries. Works like \cite{He2008adasyn,Lin2013Dys,barua2014mvmote}, on the other hand, asymmetrically oversample the minority class such that more synthetic points are generated surrounding the instances which are difficult to classify. Although these methods achieved commendable improvement on classical classifiers, they can neither be extended to deep learning techniques nor be applied to images, respectively due to the end-to-end structure of deep learning algorithms and a lack of proper notion of distance between images.

Extending GANs for semi-supervised learning, works like \cite{kumar2017semi, salimans2016improved} fused a $c$-class classifier with the discriminator by introducing an extra output line to identify the fake samples. On the other hand, \cite{springenberg2015unsupervised} proposed a $c$-class discriminator which makes uncertain predictions for fake images. Additionally, \cite{odena2017conditional} proposed a shared discriminator-cum-classifier network which makes two separate sets of predictions using two different output layers. These approaches can loosely be considered to be related to GAMO as these also incorporate a classifier into the adversarial learning scheme.

\vspace{-0.05in}

\begin{figure*}[!ht]
\vskip 0.2in
\begin{center}
\subfigure[\label{fig:sim1}]{\includegraphics[width=0.22\linewidth]{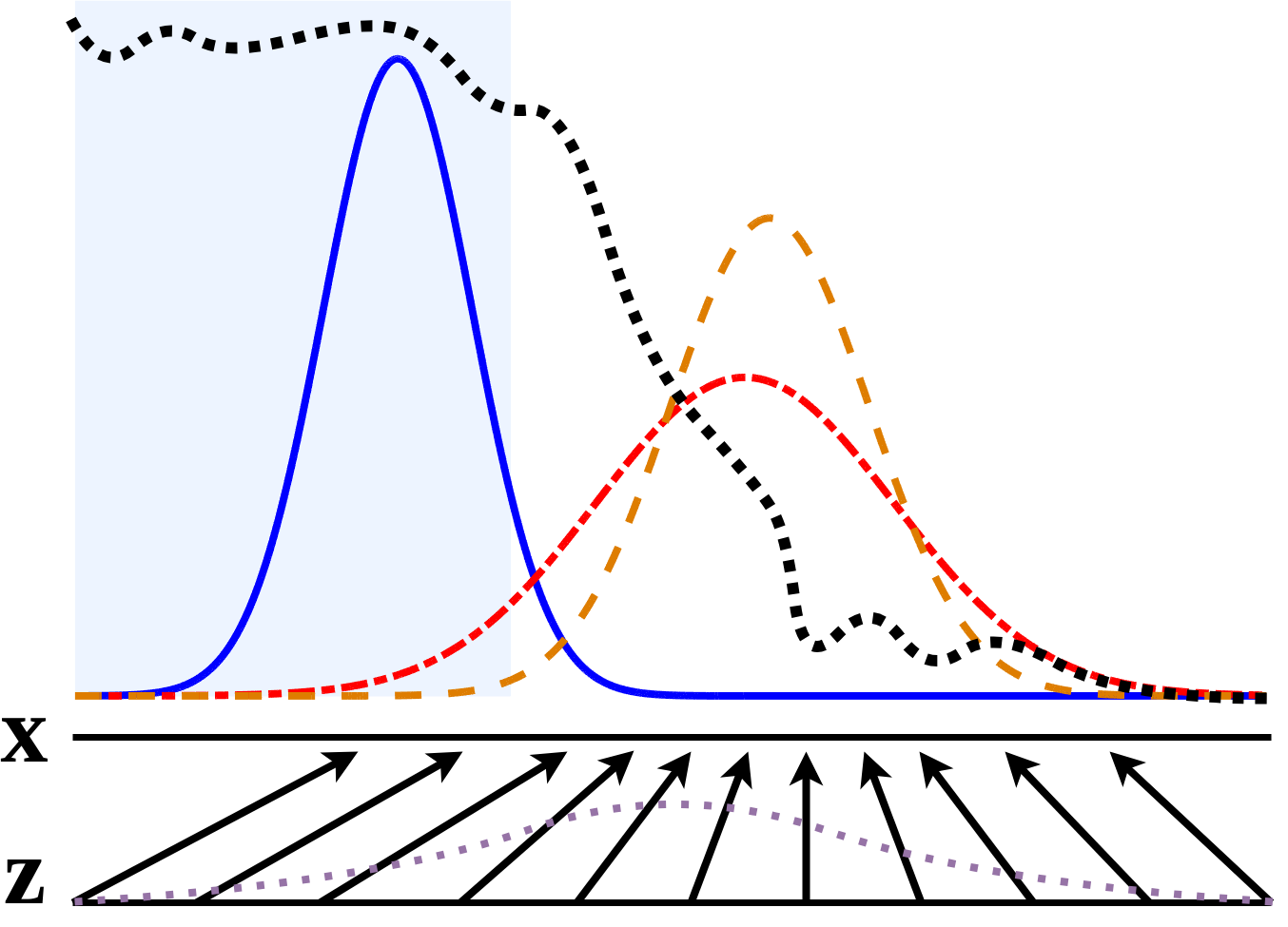}}
\subfigure[\label{fig:sim2}]{\includegraphics[width=0.22\linewidth]{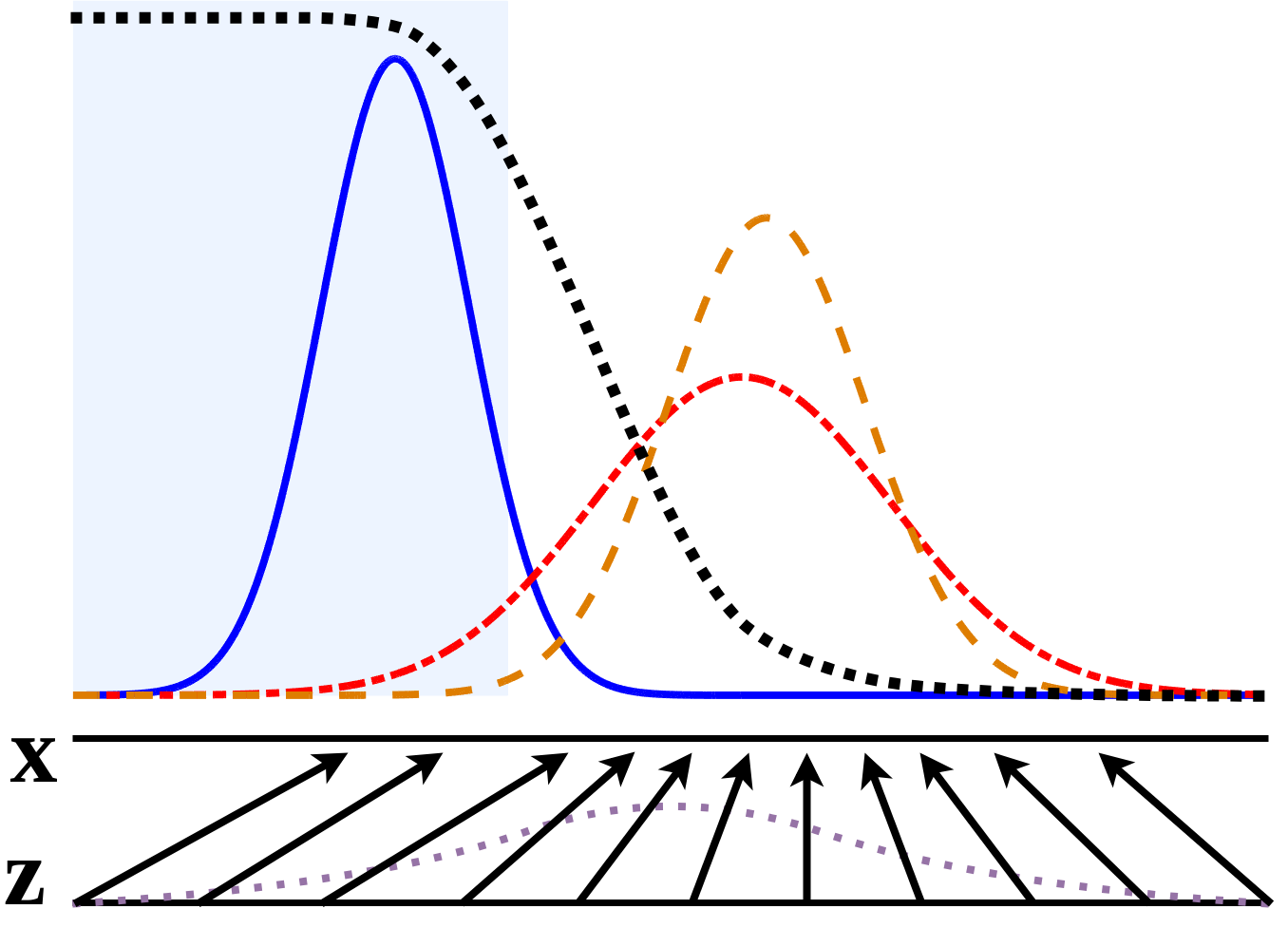}}
\subfigure[\label{fig:sim3}]{\includegraphics[width=0.22\linewidth]{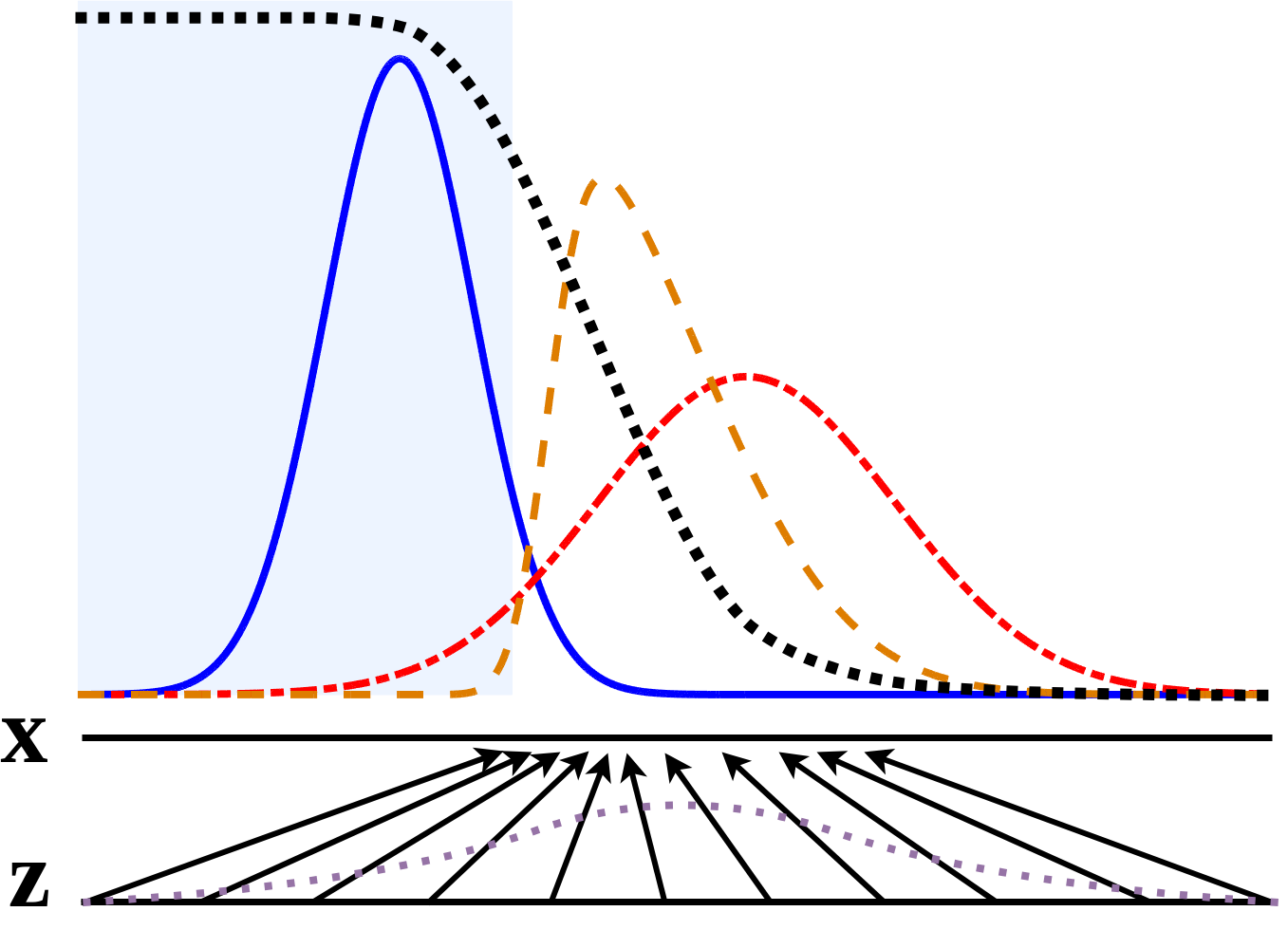}}
\subfigure[\label{fig:sim4}]{\includegraphics[width=0.22\linewidth]{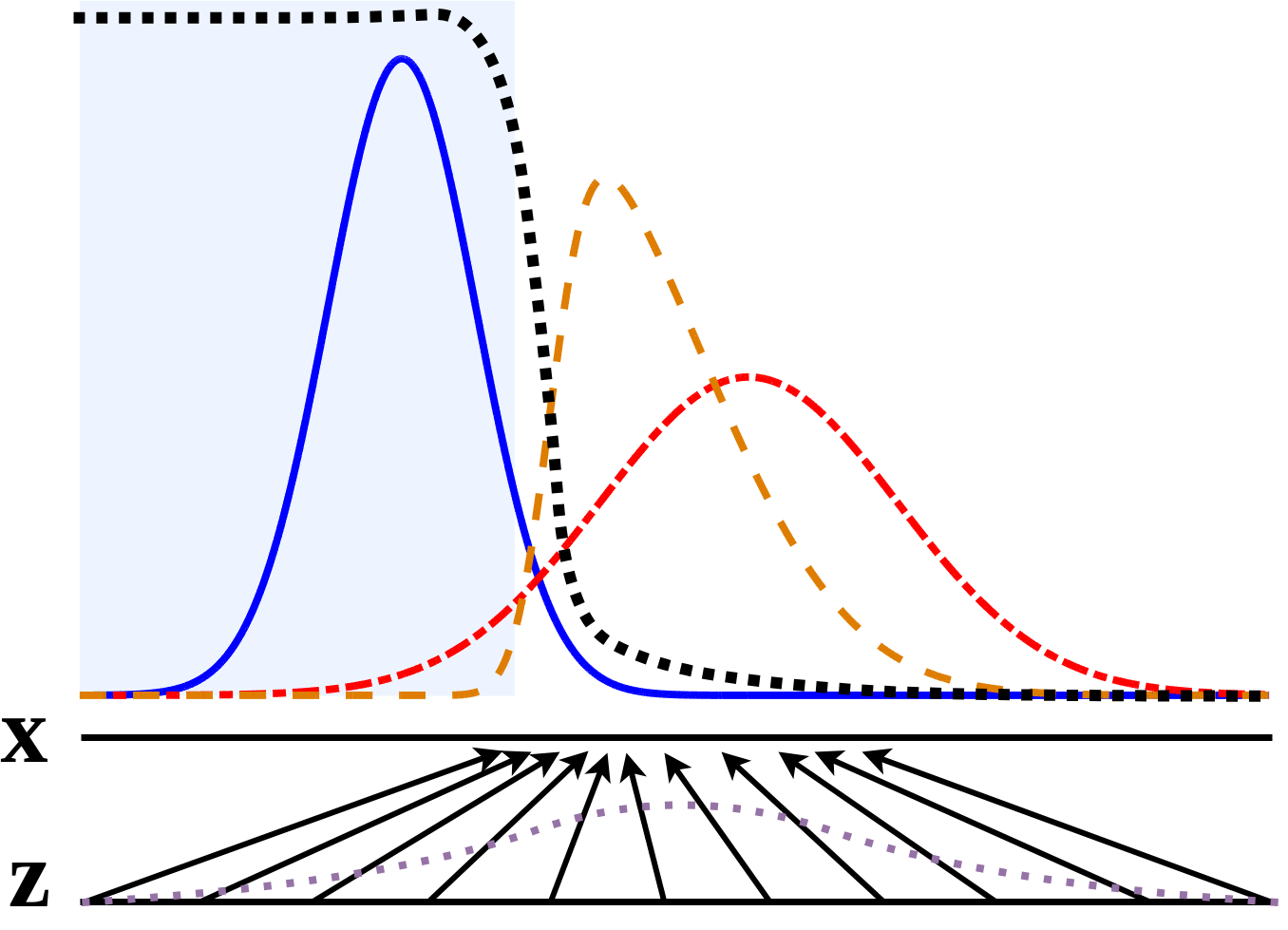}}
\caption{GAMO functions by simultaneously updating the classifier $M$ and the generator $G$. The classification function (black, dotted line) is trained to correctly classify samples from the majority class distribution $p^d_{maj}$ (blue, solid line), the real minority class distribution $p^d_{min}$ (red, dots and dashes) as well as the generated minority distribution $p^g_{min}$ (brown, dashed line). The generator, on the other hand, is trained to generate minority samples which will be misclassified by $M$. The upward arrows show how the generator learns the mapping $\mathbf{x} = G(\mathbf{z})$ from a standard normal distribution (mauve, dotted line) in the latent space to convex combinations of the real minority instances from the minority class. The ideal classification function is shown as a blue highlight in the background. (a) Let us consider an initial adversarial pair: the generated distribution $p^g_{min}$ is similar to the real distribution of the minority class $p^d_{min}$ and $M$ is an inaccurate classifier. (b) $M$ is trained to properly classify the samples from the three distributions $p^d_{maj}$, $p^d_{min}$, and $p^g_{min}$; resulting in a non-ideal trained classifier which is biased in favor of the majority class. (c) After an update to $G$, the gradient of $M$ has guided $G(\mathbf{z})$ to flow to regions that are more likely to be misclassified by $M$. (d) Thereafter, retraining $M$ results in a classifier much closer to the ideal classifier due to the increased number of minority samples near the boundary of the two classes.}
\label{fig:illustration}
\end{center}
\vskip -0.3in
\end{figure*}

\section{Proposed Method}\label{sec:method}

Let us consider a $c$-class classification problem with a training dataset $X \subset \mathbb{R}^{D}$ (of images vectorized either by flattening or by a convolutional feature extraction network $F$). Let the prior probability of the $i$-th class be $P_{i}$, where $i \in \mathcal{C}=\{1, 2, \cdots c\}$; $\mathcal{C}$ being the set of possible class labels. Without loss of generality, we consider the classes to be ordered such that $P_{1} \leq P_{2} \leq \cdots < P_{c}$. Moreover, let $X_{i}$ denote the set of all $n_{i}$ training points which belong to class $i \in \mathcal{C}$. We intend to train a classifier $M$ having $c$ output lines, where the $i$-th output $M_{i}(\mathbf{x})$ predicts the probability of any $\mathbf{x}\in X$ to be a member of the $i$-th class.

\subsection{Adversarial Oversampling}

Our method plays an adversarial game between a classifier that aims to correctly classify the data points and a generator attempting to spawn artificial points which will be misclassified by the classifier. The idea is that generating such difficult points near the fringes of the minority class(es) will help the classifier to learn class boundaries which are more robust to class imbalance. In other words, the performance of the classifier will adversarially guide the generator to generate new points at those regions where the minority class under concern is prone to misclassification. Moreover, the classifier will aid the generator to adaptively determine the concentration of artificial instances required to improve the classification performance in a region, thus relieving the user from tuning the amount of oversampling. Instead, we only need to fix the number of points to be generated to the difference between the number of points in the majority class and that of the (respective) minority class(es). 

\begin{figure*}[!t]
    \centering
    \includegraphics[width=0.97\linewidth]{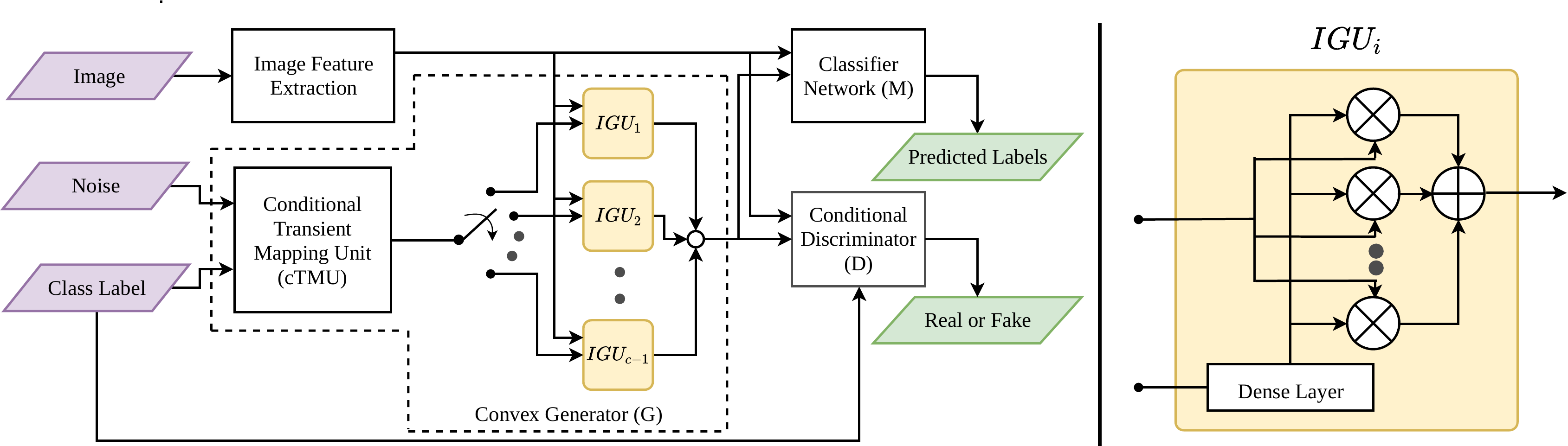}
    \caption{The GAMO model: (Left) Schematic of the GAMO framework; (Right) Illustration of an Instance Generation Unit ($IGU$). Given an image, the extracted feature vectors (either by a convolutional neural network $F$ or by flattening) are fed to the classifier network $M$ as well as the conditional discriminator $D$. $M$ predicts the class label for the input data point while $D$ distinguishes between real and fake data instances. The convex generator network $G$ is composed of a  $cTMU$, and $IGU$s corresponding to each of the $c-1$ minority classes. The $IGU_{i}$ network takes an intermediate vector generated by $cTMU$ and maps it to a set of $n_{i}$ convex weights. It then takes the set $X_{i}$ as input and generates a new sample for the $i$-th class, as the convex combination of all the $\mathbf{x}_{j} \in X_{i}$.}
    \label{fig:gamoModel}
    \vskip -0.1in
\end{figure*}

\subsection{Convex Generator}

The generator tries to generate points which will be misclassified by the classifier. Hence, if left unchecked, the generator may eventually learn to generate points which do not coincide with the distribution of the intended minority class. This may help improve the performance on the concerned minority class but will lead to high misclassification from the other classes. To prevent this from happening, we generate the new points only as convex combinations of the existing points from the minority class in question. This will restrict the generated distribution within the convex hull of the real samples from the (respective) minority class(es). Since the generator attempts to conjure points that are difficult for the classifier, the points are generated near the peripheries of the minority class(es).

Our convex generator $G$ comprises of two modules: a Conditional Transient Mapping Unit ($cTMU$) and a set of class-specific Instance Generation Units ($IGU$), which we propose to limit the model complexity. The $cTMU$ network learns a mapping $t$, conditioned on class $i$, from a $l$-dimensional latent space to an intermediate space. The $IGU_{i}$, on the other hand, learns a mapping $g_{i}$ from the $cTMU$ output space to a set of $n_{i}$ convex weights $g_{i}(t(\mathbf{z}|i)) \geq 0$, s.t. $\sum_{j=1}^{n_{i}}g_{i}(t(\mathbf{z}|i))=1$, using softmax activation. Hence, $G$ can generate a new $D$-dimensional sample for the $i$-th class as a convex combination of the data points in $X_{i}$,
\vskip -0.12in
\begin{equation}
    G(\mathbf{z}|i)=\sum_{j=1}^{n_{i}}g_{i}(t(\mathbf{z}|i))\mathbf{x}_{j},
\end{equation}
\vskip -0.04in
where $\mathbf{z}$ is a latent variable drawn from a standard normal distribution and $\mathbf{x}_{j} \in X_{i}$.

Formally, the adversarial game played by the proposed classifier-convex generator duo poses the following optimization problem, when cross entropy loss is considered: 
\vskip -0.2in
\begingroup
\allowdisplaybreaks
\begin{align}
    \label{eqn:objective}  
    & \quad \quad \quad \quad \quad \; \min_{G} \max_{M} J(G, M)= \sum_{i \in \mathcal{C}} J_{i}, \\
    &\text{where \;} J_{i}=(J_{i1}+J_{i2}+J_{i3}+J_{i4}), \nonumber \\
    &J_{i1}=P_{i}\mathbb{E}_{\mathbf{x} \sim p^{d}_{i}} [\log M_{i}(\mathbf{x})], \nonumber \\
    &J_{i2}=\sum_{j \in \mathcal{C}\setminus \{i\}}P_{j}\mathbb{E}_{\mathbf{x} \sim p^{d}_{j}} [\log (1-M_{i}(\mathbf{x}))], \nonumber \\ 
    &J_{i3}=(P_{c}-P_{i})\mathbb{E}_{G(\mathbf{z}|i) \sim p^{g}_{i}} [\log M_{i}(G(\mathbf{z}|i))], \text{and}, \nonumber \\
    &J_{i4}=\sum_{j \in \mathcal{C}\setminus \{i\}}(P_{c}-P_{j})\mathbb{E}_{G(\mathbf{z}|j) \sim p^{g}_{j}} [\log (1-M_{i}(G(\mathbf{z}|j)))], \nonumber
\end{align}
\endgroup
\vskip -0.05in
while $p^{d}_{i}$ and $p^{g}_{i}$ respectively denote the real and generated class conditional probability distributions of the $i$-th class.

The two-player minimax game formalized in (\ref{eqn:objective}) is played between a classifier $M$ and a generator $G$. $M$ attempts to correctly classify all real as well as generated points belonging to all the classes. Whereas, $G$ strives to generate sample(s) which have a high probability of being classified by $M$ into all other classes. To demonstrate how such an adversarial game can aid $M$ to learn a better class boundary, we illustrate its chronological progression in a more explanatory manner in Figure \ref{fig:illustration}. In Theorem \ref{thm:JS}, we show that the optimization problem in (\ref{eqn:objective}) is equivalent to minimizing a sum of the Jensen-Shannon divergences.
\vspace{-0.02in}
\begin{theorem}\label{thm:JS}
Optimizing the objective function $J$ is equivalent to the problem of minimizing the following summation of Jensen-Shannon divergences:
\begin{equation*}
    \sum_{i=1}^{c}JS\Big( \big( P_{i}p_{i}^{d}+(P_{c}-P_{i})p_{i}^{g} \big) \Big| \Big|
    \sum_{\substack{j \neq i \\ j=1}}^{c} \big( P_{j}p_{j}^{d}+(P_{c}-P_{j})p_{j}^{g} \big) \Big) 
\end{equation*}
\begin{proof}
See the supplementary document.
\end{proof}
\end{theorem}

The behavior of the proposed approach can be understood by interpreting Theorem \ref{thm:JS}. The optimization problem aims to bring the generated distribution, for a particular class, closer to the generated as well as real distributions for all other classes. Since the real distributions are static for a fixed dataset, the optimization problem in Theorem \ref{thm:JS} essentially attempts to move the generated distributions for each class closer to the real distributions for all other classes. This is likely to result in the generation of ample points near the peripheries, which are critical to combating class imbalance. While doing so, the generated distributions for all classes also strive to come closer to each other. However, the generated distributions for the different classes do not generally collapses upon each other, being constrained to remain within the convex hulls of the respective classes. 

\subsection{Additional Discriminator}

While the generator only generates points within the convex hull of the samples from the minority class(es), the generated points may still be placed at locations within the convex hull which do not correspond to the distribution of the intended class (recall Figure \ref{fig:exampleGamoNoDis}). This is likely to happen if the intended minority class(es) are non-convex in shape. Moreover, we know from Theorem \ref{thm:JS} that the generated distributions for different minority classes may come close to each other if the respective convex hulls overlap. To solve this problem, we introduce an additional conditional discriminator which ensures that the generated points do not fall outside the actual distribution of the intended minority class(es). Thus, the final adversarial learning system proposed by us consists of three players, viz. a multi-class classifier $M$, a conditional discriminator $D$ which given a class aims to distinguish between real and generated points, and a convex generator $G$ that attempts to generate points which, in addition to being difficult for $M$ to correctly classify, are also mistaken by $D$ to be real points sampled from the given dataset. The resulting three-player minimax game is formally presented in (\ref{eqn:objDis}).
\vspace{-0.1in}
\begin{algorithm*}[!ht]
  \caption{\small Generative Adversarial Minority Oversampling (GAMO)}
  \label{alg:gamo}
  \footnotesize
  {\bfseries Input:} $X$: training set, $l$: latent dimension, $b$: minibatch size, $u$, $v$: (hyperparameters, set to $\lceil \frac{n}{b} \rceil$ in our implementation). \\
  {\bfseries Output:} A trained classification network $M$. \\
  {\bfseries Note:} For flattened images there is no need to train $F$, i.e., $F(X)$ can be replaced by $X$.
  \let \oldnoalign \noalign
  \let \noalign \relax
  \midrule
  \let \noalign \oldnoalign
  \begin{algorithmic}[1]
  \WHILE{not converged} 
        \FOR{$u$ steps}
            \STATE Sample $B_{d}=\{\mathbf{x}_{1}, \mathbf{x}_{2}, \cdots \mathbf{x}_{b}\}$ from $X$, with corresponding class labels $Y_{d}$. 
            \STATE Update $F$ by gradient descent on $(M(F(B_{d})), Y_{d})$ keeping $M$ fixed.
        \ENDFOR
        \FOR{$v$ steps}
            \STATE Sample $B_{d}=\{\mathbf{x}_{1}, \mathbf{x}_{2}, \cdots \mathbf{x}_{b}\}$ from $X$, with corresponding class labels $Y_{d}$.
            \STATE Sample $B_{n}=\{\mathbf{z}_{1}, \mathbf{z}_{2}, \cdots \mathbf{z}_{b}\}$ from $l$ dimensional standard normal distribution.
            \STATE Update $M$ and $D$ by respective gradient descent on $(M(F(B_{d})), Y_{d})$ and $(D(F(B_{d})|Y_{d}), \mathbf{1})$, keeping $F$ fixed.
            \STATE Generate labels $Y_{n}$ by assigning each $\mathbf{z}_{j} \in B_{n}$ to one of the $c-1$ minority classes, with probability $\propto$ $(P_{c}-P_{i})$; $\forall i \in \mathcal{C}\setminus\{c\}$.
            \STATE Update $M$ and $D$ by respective gradient descent on $(M(G(B_{n}|Y_{n})), Y_{n})$ and $(D(G(B_{n}|Y_{n})|Y_{n}), \mathbf{0})$, keeping $G$ fixed.
            \STATE Sample $B_{g}=\{\mathbf{z}_{1}, \mathbf{z}_{2}, \cdots \mathbf{z}_{b}\}$ from $l$ dimensional standard normal distribution.
            \STATE Generate labels $Y_{g}$ by assigning each $\mathbf{z}_{j} \in B_{g}$ to any of the $c-1$ minority classes with equal probability. Take ones' complement of $Y_{g}$ as $\overline{Y_{g}}$.
            \STATE Update $G$ by gradient descent on $(M(G(B_{g}|Y_{g})), \overline{Y_{g}})$ keeping $M$ fixed.
            \STATE Update $G$ by gradient descent on $(D(G(B_{g}|Y_{g})|Y_{g}), \mathbf{1})$ keeping $D$ fixed. 
        \ENDFOR
    \ENDWHILE
\end{algorithmic}
\end{algorithm*}
\begingroup
\allowdisplaybreaks
\begin{align}
\label{eqn:objDis}
    & \min_{G} \max_{M} \max_{D} Q(G, M, D)= \sum_{i \in \mathcal{C}} Q_{i}, \\
    & \text{where, } Q_{i}=(J_{i1}+J_{i2}+J_{i3}+J_{i4}+Q_{i1}+Q_{i2}), \nonumber \\
    & Q_{i1}=P_{i}\mathbb{E}_{\mathbf{x} \sim p^{d}_{i}} [\log D(\mathbf{x}|i)], \text{and}, \nonumber \\
    & Q_{i2}=(P_{c}-P_{i})\mathbb{E}_{G(\mathbf{z}|i) \sim p^{g}_{i}} [\log (1-D(G(\mathbf{z}|i)|i))]. \nonumber
\end{align}
\endgroup
\vspace{-0.1in}

\vspace{-0.1in}
\subsection{Least-Square Formulation}
Mao et al. \cite{mao2017least} showed that replacing the popular cross entropy loss in GAN with least square loss can not only produce better quality images but also can prevent the vanishing gradient problem to a greater extent. Therefore, we also propose a variant of GAMO using the least square loss, which poses the following optimization problem:
\vspace{-0.2in}

\begingroup
\allowdisplaybreaks
\begin{align}
    & \min_{M} L_{M}=\sum_{i \in \mathcal{C}} (L_{i1}+L_{i2}+L_{i3}+L_{i4}), \label{eqn:lsObjM} \\
    & \min_{D} L_{D}=\sum_{i \in \mathcal{C}} (L_{i5}+L_{i6}) \label{eqn:lsObjD}, \\
    & \min_{G} L_{G}=\sum_{i \in \mathcal{C}\setminus \{c\}} (L_{i7}+L_{i8}+L_{i9}) \label{eqn:lsObjG}, \\
    & \text{where, } L_{i1}=P_{i}\mathbb{E}_{\mathbf{x} \sim p^{d}_{i}} [(1-M_{i}(\mathbf{x}))^{2}], \nonumber \\
    & L_{i2}=\sum_{j \in \mathcal{C}\setminus \{i\}}P_{j}\mathbb{E}_{\mathbf{x} \sim p^{d}_{j}} [(M_{i}(\mathbf{x}))^{2}], \nonumber \\ 
    & L_{i3}=(P_{c}-P_{i})\mathbb{E}_{G(\mathbf{z}|i) \sim p^{g}_{i}} [(1-M_{i}(G(\mathbf{z}|i)))^{2}], \nonumber \\
    & L_{i4}=\sum_{j \in \mathcal{C}\setminus \{i\}}(P_{c}-P_{j})\mathbb{E}_{G(\mathbf{z}|j) \sim p^{g}_{j}} [(M_{i}(G(\mathbf{z}|j)))^{2}], \nonumber \\
    & L_{i5}=P_{i}\mathbb{E}_{\mathbf{x} \sim p^{d}_{i}} [(1-D(\mathbf{x}|i))^{2}], \nonumber \\
    & L_{i6}=(P_{c}-P_{i})\mathbb{E}_{G(\mathbf{z}|i) \sim p^{g}_{i}} [(D(G(\mathbf{z}|i)|i))^{2}], \nonumber \\
    & L_{i7}=\mathbb{E}_{G(\mathbf{z}|i) \sim p^{g}_{i}} [(M_{i}(G(\mathbf{z}|i)))^{2}], \nonumber \\
    & L_{i8}=\sum_{j \in \mathcal{C}\setminus \{i, c\}}\mathbb{E}_{G(\mathbf{z}|j) \sim p^{g}_{j}} [(1-M_{i}(G(\mathbf{z}|j)))^{2}], \text{and}, \nonumber \\
    & L_{i9}=\mathbb{E}_{G(\mathbf{z}|i) \sim p^{g}_{i}} [(1-D(G(\mathbf{z}|i)|i))^{2}]. \nonumber 
\end{align}
\endgroup

\subsection{Putting it all together}
The model for the GAMO framework is detailed in Figure \ref{fig:gamoModel}, while the complete algorithm is described in Algorithm \ref{alg:gamo}. To ensure an unbiased training for $M$ and $D$ we generate artificial points for the $i$-th class with probability $(P_{c} - P_{i})$ to compensate for the effect of imbalance. On the other hand, to also ensure unbiased training for $G$ we use samples from all classes with equal probability.


\section{Experiments}\label{sec:exp}
We evaluate the performance of a classifier in terms of two indices which are not biased toward any particular class \cite{Sokolova2009}, namely Average Class Specific Accuracy (ACSA) \cite{huang2016learning, wang2017meta} and Geometric Mean (GM) \cite{kubat1997, branco2016}. All our experiments have been repeated 10 times to mitigate any bias generated due to randomization and the means and standard deviations of the index values are reported. Codes for the proposed methods are available at \url{https://github.com/SankhaSubhra/GAMO}.

\begin{table*}[!ht]
    \centering
    \caption{Comparison of classification performance of CE and LS variants of classifiers on MNIST and Fashion-MNIST datasets.}
    \label{tab:mnistFmnistRes}
    \vskip 0.15in
    \scriptsize
    \begin{tabular}{ccccccccc} 
    \toprule
    & \multicolumn{4}{c}{MNIST} & \multicolumn{4}{c}{Fashion-MNIST} \\ \cmidrule{2-9}
    Algorithm & \multicolumn{2}{c}{CE} & \multicolumn{2}{c}{LS} & \multicolumn{2}{c}{CE} & \multicolumn{2}{c}{LS} \\ \cmidrule{2-9}
    & ACSA & GM & ACSA & GM & ACSA & GM & ACSA & GM \\ 
    \midrule
    Baseline CN & 0.88$\pm$0.01 & 0.87$\pm$0.02 & 0.88$\pm$0.01 & 0.86$\pm$0.01 & 0.82$\pm$0.01 & 0.80$\pm$0.01 & 0.81$\pm$0.01 & 0.79$\pm$0.01 \\ 
    SMOTE+CN & 0.88$\pm$0.02 & 0.87$\pm$0.03 & 0.89$\pm$0.01 & 0.89$\pm$0.01 & - & - & - & - \\
    Augment+CN & - & - & - & - & 0.82$\pm$0.01 & 0.78$\pm$0.01 & 0.82$\pm$0.01 & 0.78$\pm$0.01 \\
    DOS & - & - & - & - & 0.82$\pm$0.01 & 0.79$\pm$0.01 & 0.81$\pm$0.01 & 0.79$\pm$0.02 \\
    (cGAN/cDCGAN)+CN & 0.88$\pm$0.01 & 0.87$\pm$0.01 & 0.89$\pm$0.01 & 0.88$\pm$0.01 & 0.81$\pm$0.02 & 0.78$\pm$0.01 & 0.82$\pm$0.01 & 0.80$\pm$0.01 \\
    cG+CN & 0.86$\pm$0.03 & 0.85$\pm$0.02 & 0.86$\pm$0.03 & 0.85$\pm$0.03 & 0.79$\pm$0.02 & 0.77$\pm$0.02 & 0.80$\pm$0.01 & 0.77$\pm$0.02 \\
    cG+D+CN & 0.85$\pm$0.02 & 0.83$\pm$0.01 & 0.85$\pm$0.02 & 0.82$\pm$0.02 & 0.79$\pm$0.02 & 0.78$\pm$0.01 & 0.79$\pm$0.01 & 0.78$\pm$0.02 \\
    GAMO\textbackslash D (Ours) & 0.87$\pm$0.01 & 0.86$\pm$0.01 & 0.88$\pm$0.01 & 0.87$\pm$0.01 & 0.81$\pm$0.01 & 0.80$\pm$0.01 & 0.82$\pm$0.01 & 0.80$\pm$0.01 \\
    GAMO (Ours) & 0.89$\pm$0.01 & 0.88$\pm$0.01 & \textbf{0.91$\pm$0.01} & \textbf{0.90$\pm$0.01} & 0.82$\pm$0.01 & 0.80$\pm$0.01 & \textbf{0.83$\pm$0.01} & \textbf{0.81$\pm$0.01} \\ 
    \bottomrule
    \end{tabular}
\end{table*}

We have used a collection of 7 image datasets for our experiments, namely MNIST \cite{lecun1998mnist}, Fashion-MNIST \cite{xiao2017fashion}, CIFAR10 \cite{krizhevsky2009cifar}, SVHN \cite{netzer2011svhn}, LSUN \cite{yu2015lsun} and SUN397 \cite{xiao2010sun}. All the chosen datasets except SUN397 are not significantly imbalanced in nature, therefore we have created their imbalanced variants by randomly selecting a disparate number of samples from the different classes. Further, for all the datasets except SUN397, 100 points are selected from each class to form the test set. In the case of SUN397 (50 classes of which are used for our experiments) 20 points from each class are kept aside for testing.

We refrain from using pre-trained networks for our experiments as the pre-learned weights may not reflect the imbalance between the classes. We, instead, train the models from scratch to emulate real-world situations where the data is imbalanced and there is no pre-trained network available that can be used as an appropriate starting point. We have obtained the optimal architectures and hyperparameters for each contending method in Section \ref{sec:exp}-\ref{sec:imgGen} using a grid search (see supplementary document).

\begin{figure}[!t]
\centering
\includegraphics[width=0.98\linewidth]{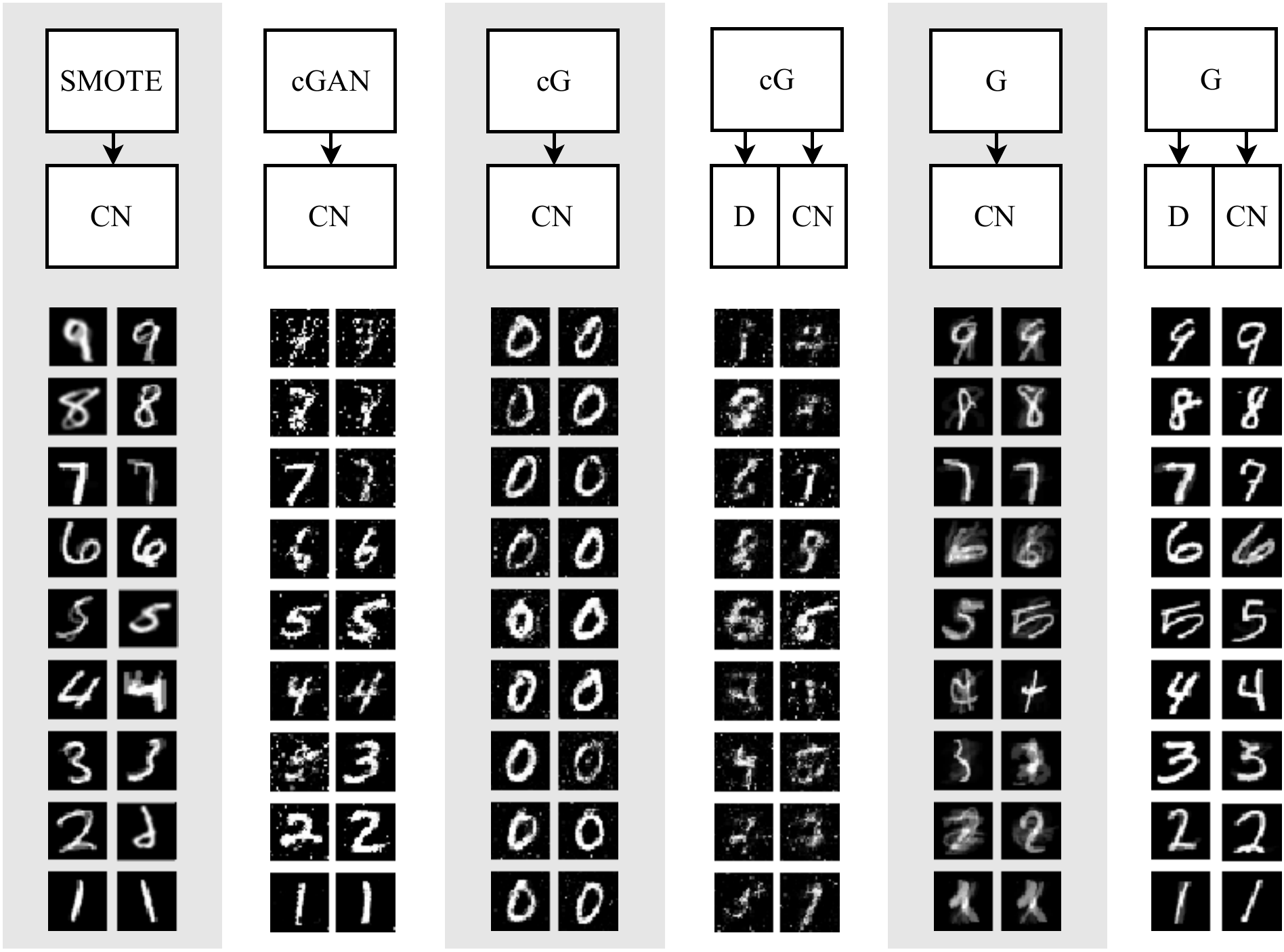}
\caption{Ablation study on the MNIST dataset: SMOTE generates artificial samples from the minority class(es) as convex combinations of pairs of neighbors from the respective class(es). The oversampled dataset is then classified using a classifier network CN. SMOTE sometimes generates unrealistic ``out-of-distribution" samples which are combinations of visually disparate images that happen to be Euclidean neighbors in the flattened image space. Using cGAN for generating new samples results in realistic images only from the more abundant minority classes. Training only a conditional Generator cG adversarially against CN, to generate images which will be misclassified by CN, results in new samples which all resemble the majority class `0'. Introducing a discriminator D (to ensure that cG adheres to class distributions) into the mix results in new samples which are somewhat in keeping with the class identities, but still unrealistic in appearance. Employing our proposed convex generator G to generate new samples by training it adversarially with CN (the GAMO\textbackslash D formulation) results in samples which are in keeping with the class identities, but often ``out-of-distribution" as the classes are non-convex. Finally, introducing D into this framework results in the complete GAMO model which can generate realistic samples which are also in keeping with the class identities.}
\label{fig:mnist_resmod}
\vskip -0.35in
\end{figure}

\begin{table}[!ht]
    \centering
    \caption{Comparison of classification performance on CIFAR10 and SVHN datasets.}
    \label{tab:cifarSvhnRes}
    \vskip 0.15in
    \scriptsize
    \begin{tabular}{cccc} 
    \toprule
    Dataset & Algorithm & ACSA & GM \\ 
    \midrule
    \multirow{5}{*}{CIFAR10} & Baseline CN & 0.45$\pm$0.01 & 0.37$\pm$0.01 \\ 
    & Augment+CN & 0.47$\pm$0.01 & 0.39$\pm$0.02 \\
    & cDCGAN+CN & 0.42$\pm$0.02 & 0.32$\pm$0.03 \\
    & DOS & 0.46$\pm$0.02 & 0.37$\pm$0.01 \\
    & GAMO\textbackslash D (Ours) & 0.47$\pm$0.01 & 0.40$\pm$0.01 \\
    & GAMO (Ours) & \textbf{0.49$\pm$0.01} & \textbf{0.43$\pm$0.02} \\
    \midrule
    \multirow{5}{*}{SVHN} & Baseline CN & 0.74$\pm$0.01 & 0.73$\pm$0.01 \\ 
    & Augment+CN & 0.69$\pm$0.01 & 0.63$\pm$0.01 \\
    & cDCGAN+CN & 0.69$\pm$0.01 & 0.66$\pm$0.02 \\
    & DOS & 0.71$\pm$0.02 & 0.68$\pm$0.01 \\
    & GAMO\textbackslash D (Ours) & 0.75$\pm$0.01 & 0.75$\pm$0.02 \\
    & GAMO (Ours) & \textbf{0.76$\pm$0.01} & \textbf{0.75$\pm$0.02} \\ 
    \bottomrule
    \end{tabular}
    \vskip -0.2in
\end{table}

\begin{table*}[!ht]
    \centering
    \caption{Comparison of classification performance with increased number of training instances on CelebA and LSUN datasets.}
    \label{tab:celebaLsunRes}
    \vskip 0.15in
    \scriptsize
    \begin{tabular}{ccccccccc} 
    \toprule
    & \multicolumn{4}{c}{CelebA-Small} & \multicolumn{4}{c}{CelebA-Large} \\ \cmidrule{2-9}
    Algorithm & \multicolumn{2}{c}{During Training} & \multicolumn{2}{c}{During Testing} & \multicolumn{2}{c}{During Training} & \multicolumn{2}{c}{During Testing} \\ \cmidrule{2-9}
    & ACSA & GM & ACSA & GM & ACSA & GM & ACSA & GM \\ 
    \midrule
    Baseline CN & 0.91$\pm$0.01 & 0.91$\pm$0.01 & 0.59$\pm$0.01 & 0.45$\pm$0.04 & 0.93$\pm$0.01 & 0.92$\pm$0.01 & 0.71$\pm$0.01 & 0.60$\pm$0.03 \\
    Augment+CN & 0.74$\pm$0.06 & 0.70$\pm$0.09 & 0.62$\pm$0.05 & 0.47$\pm$0.08 & 0.82$\pm$0.01 & 0.79$\pm$0.01 & 0.72$\pm$0.01 & 0.66$\pm$0.02 \\
    cDCGAN+CN & 0.86$\pm$0.01 & 0.84$\pm$0.01 & 0.59$\pm$0.01 & 0.36$\pm$0.02 & 0.87$\pm$0.01 & 0.86$\pm$0.01 & 0.67$\pm$0.01 & 0.58$\pm$0.02 \\
    DOS & 0.82$\pm$0.03 & 0.80$\pm$0.02 & 0.61$\pm$0.01 & 0.48$\pm$0.02 & 0.84$\pm$0.01 & 0.83$\pm$0.02 & 0.72$\pm$0.01 & 0.64$\pm$0.02 \\
    GAMO (Ours) & 0.92$\pm$0.01 & 0.91$\pm$0.01 & \textbf{0.66$\pm$0.01} & \textbf{0.54$\pm$0.02} & 0.91$\pm$0.01 & 0.91$\pm$0.01 & \textbf{0.75$\pm$0.01} & \textbf{0.70$\pm$0.02} \\
    \cmidrule{2-9}
    & \multicolumn{4}{c}{LSUN-Small} & \multicolumn{4}{c}{LSUN-Large} \\ \cmidrule{2-9}
    & \multicolumn{2}{c}{During Training} & \multicolumn{2}{c}{During Testing} & \multicolumn{2}{c}{During Training} & \multicolumn{2}{c}{During Testing} \\ \cmidrule{2-9}
    & ACSA & GM & ACSA & GM & ACSA & GM & ACSA & GM \\ 
    \cmidrule{2-9}
    Baseline CN & 0.90$\pm$0.01 & 0.89$\pm$0.01 & 0.50$\pm$0.01 & 0.28$\pm$0.05 & 0.87$\pm$0.01 & 0.87$\pm$0.01 & 0.61$\pm$0.02 & 0.54$\pm$0.03 \\ 
    Augment+CN & 0.67$\pm$0.06 & 0.64$\pm$0.09 & 0.54$\pm$0.03 & 0.45$\pm$0.07 & 0.70$\pm$0.03 & 0.65$\pm$0.03 & 0.64$\pm$0.02 & 0.58$\pm$0.03 \\
    cDCGAN+CN & 0.80$\pm$0.02 & 0.79$\pm$0.02 & 0.53$\pm$0.02 & 0.43$\pm$0.03 & 0.81$\pm$0.02 & 0.80$\pm$0.02 & 0.60$\pm$0.02 & 0.53$\pm$0.03 \\
    DOS & 0.78$\pm$0.03 & 0.76$\pm$0.02 & 0.54$\pm$0.02 & 0.44$\pm$0.02 & 0.79$\pm$0.02 & 0.77$\pm$0.02 & 0.63$\pm$0.02 & 0.61$\pm$0.03 \\
    GAMO (Ours) & 0.93$\pm$0.01 & 0.93$\pm$0.01 & \textbf{0.57$\pm$0.01} & \textbf{0.50$\pm$0.02} & 0.80$\pm$0.01 & 0.80$\pm$0.01 & \textbf{0.70$\pm$0.02} & \textbf{0.68$\pm$0.03} \\
    \bottomrule
    \end{tabular}
    \vskip -0.1in
\end{table*}

\subsection{MNIST and Fashion-MNIST}

The experiments in this section are conducted using imbalanced subsets of the MNIST and Fashion-MNIST datasets. In case of both the datasets, we have sampled $\{4000, 2000, 1000, 750, 500, 350, 200, 100, 60, 40\}$ points from classes in order of their index. Thus, the datasets have an Imbalance Ratio (IR: ratio of the number of representatives from the largest class to that of the smallest class) of 100. We begin by establishing the effectiveness of our proposed framework. We also compare between the two variants of GAMO which use Cross Entropy (CE) and Least Square (LS) losses, respectively.  

We undertake an ablation study on MNIST using flattened images to facilitate straightforward visualization of the oversampled instances. Convolutional features are used for Fashion-MNIST. For MNIST, we have compared GAMO, against baseline classifier network (CN), SMOTE+CN (training set is oversampled by SMOTE), cGAN+CN (training set oversampled using cGAN, which is then used to train CN), and also traced the evolution of the philosophy behind GAMO, through cG+CN (conditional generator cG adversarially trained against CN, in contrast to cGAN+CN where CN does not play any part in training cGAN), cG+D+CN (cG+CN network coupled with a discriminator D), and GAMO\textbackslash D (GAMO without a discriminator) on the MNIST dataset. SMOTE+CN and cGAN+CN are respectively replaced by Augment+CN (data augmentation is used to create new images for balancing the training sets), and cDCGAN+CN (oversampled using conditional deep convolutional GAN) for Fashion-MNIST. GAMO is also compared with DOS on Fashion-MNIST.

The ablation study is shown visually in Figure \ref{fig:mnist_resmod} and the results for both datasets are tabulated in Table \ref{tab:mnistFmnistRes}. Overall, GAMO is observed to perform better than all other methods on both datasets. Interestingly, GAMO\textbackslash D performs much worse than GAMO on MNIST but improves significantly on Fashion-MNIST. This may be due to the fact that the convolutional feature extraction for Fashion-MNIST results in distributed representations where the classes are almost convex with little overlap between classes, enabling the convex generator to always generate data points which reside inside the class distributions. 

Since we observe from Table \ref{tab:mnistFmnistRes} that the LS variants of the classifiers mostly perform better than their CE based counterparts (which according to \cite{mao2017least} is contributed by the more stable and better decision boundary learned in LS), all the experiments in the subsequent sections are reported using the LS formulation for all the contending algorithms. 

\vspace{-0.05in}
\subsection{CIFAR10 and SVHN}
In case of CIFAR10 and SVHN the classes are subsampled (4500, 2000, 1000, 800, 600, 500, 400, 250, 150, and 80 points are selected in order of the class labels) to achieve an IR of 56.25. From Table \ref{tab:cifarSvhnRes} we can see that GAMO performs better than others on both of these datasets, closely followed by GAMO\textbackslash D, further confirming the additional advantage of convolutional feature extraction in the GAMO framework. Interestingly, Augment+CN performs much worse than the other methods on the SVHN dataset. This may be due to the nature of the images in the SVHN dataset, which may contain multiple digits. In such cases, attempting to augment the images may result in a shift of focus from one digit to its adjacent digit, giving rise to a discrepancy with the class labels. 

\vspace{-0.07in}

\begin{table}[!ht]
    \centering
    \caption{Comparison of classification performance on SUN397.}
    \label{tab:sun50Res}
    \vskip 0.15in
    \scriptsize
    \begin{tabular}{ccc} 
    \toprule
    Algorithm & ACSA & GM \\
    \midrule
    Baseline CN & 0.26$\pm$0.04 & 0.19$\pm$0.05 \\
    Augment+CN & 0.30$\pm$0.04 & 0.21$\pm$0.04\\
    cDCGAN+CN & 0.20$\pm$0.05 & 0.00$\pm$0.00 \\
    DOS & 0.28$\pm$0.04 & 0.20$\pm$0.05 \\
    GAMO (Ours) & \textbf{0.32$\pm$0.04} & \textbf{0.24$\pm$0.03} \\
    \bottomrule
    \end{tabular}
    \vskip -0.2in
\end{table}

\subsection{CelebA and LSUN}
The experiment on CelebA and LSUN are undertaken to evaluate the performance of GAMO on images of higher resolution, as well as to assess the effects of an increase in the number of instances from the different classes. In case of CelebA the images are scaled to $64 \times 64$ size, while for LSUN the same is done on a central patch of resolution $224 \times 224$ extracted from each image. In the case of CelebA we have created two 5 class datasets by selecting samples from non-overlapping classes of hair colors, namely \emph{blonde, black, bald, brown}, and \emph{gray}. The first dataset is the smaller one (having 15000, 1500, 750, 300, and 150 points in the respective classes) with an IR of 100, while the second one is larger (having 28000, 4000, 3000, 1500, and 750 points in the respective classes) with an IR of 37.33. Similarly, in the case of LSUN we select 5 classes namely \emph{classroom, church outdoor, conference room, dining room}, and \emph{tower}, and two datasets are created. The smaller one (with 15000, 1500, 750, 300, and 150 points from the respective classes) has an IR of 100, while the larger one (with 50000, 5000, 3000, 1500, and 750 points) has an IR of 66.67. 

In Table \ref{tab:celebaLsunRes}, we present the ACSA and GM over both the training and test set for the small and large variants of the two datasets. We can observe that all the algorithms manage to close the gap between their respective training and testing performances as the size of the dataset increases. Moreover, while Augment+CN seems to have the lowest tendency to overfit (smallest difference between training and testing performances), GAMO exhibits a greater ability to retain good performance on the test dataset.

\begin{figure}[!ht]
\vskip 0.2in
\begin{center}
\subfigure[\label{fig:gamo2pix}]{\includegraphics[width=0.9\linewidth]{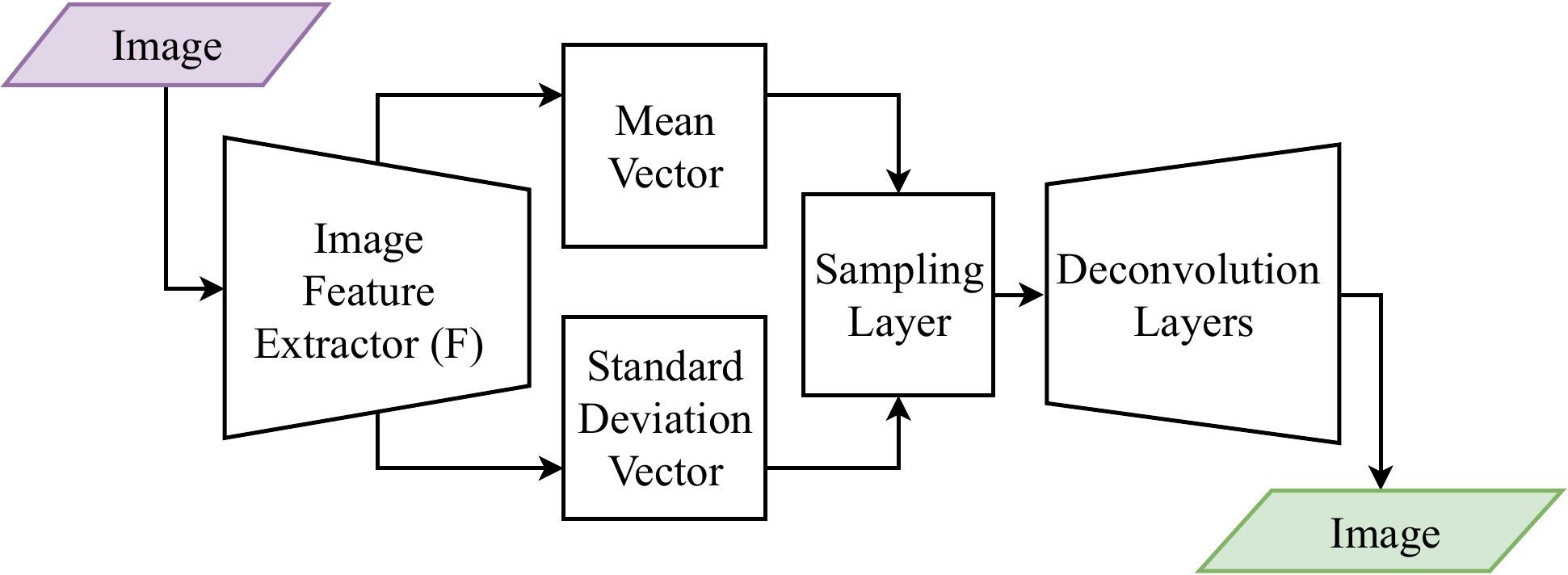}}
\subfigure[\label{fig:cdcganim}]{\includegraphics[width=0.73\linewidth]{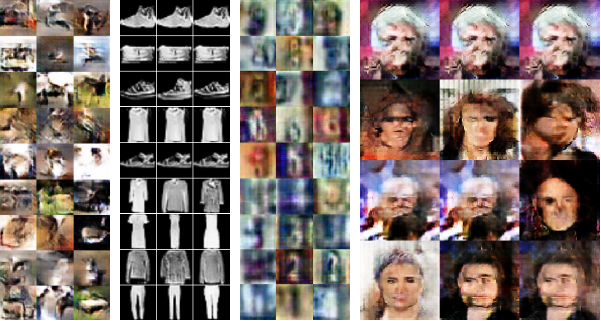}}
\subfigure[\label{fig:gamoim}]{\includegraphics[width=0.73\linewidth]{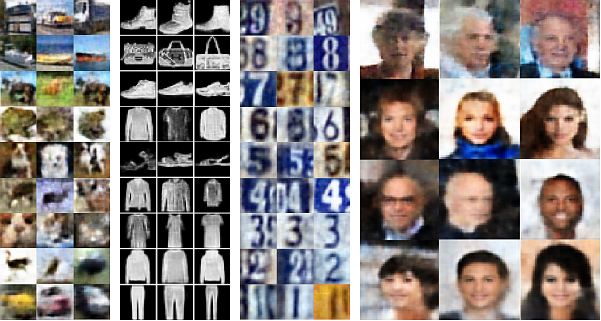}}
\caption{(a) GAMO2pix network. (b)-(c) Comparison of images respectively generated by cDCGAN, and GAMO2pix for (left to right) CIFAR10, Fashion-MNIST, SVHN, and CelebA-Small.}
\label{fig:gamoPics}
\end{center}
\vskip -0.3in
\end{figure}
\subsection{SUN397}
We have randomly selected 50 classes from SUN397 to construct a dataset containing $64 \times 64$ sized images (depending on the image size either a $512 \times 512$ or a $224 \times 224$ center patch is extracted, which is then scaled to $64 \times 64$) with an IR of 14.21. The experiment on SUN397 is performed to evaluate the performance of GAMO over a large number of classes. A scrutiny of the result tabulated in Table \ref{tab:sun50Res} reveals that despite all four contending techniques being severely affected by the complexity of the classes and the scarcity of data samples from many of the classes, GAMO is able to retain overall better performance than its competitors. 
\vspace{-0.05in}

\section{GAMO2pix}\label{sec:imgGen}
GAMO results ultimately in a classifier trained to properly classify samples from all the classes. However, some application may require that actual samples be generated by oversampling to form an artificially balanced dataset. While GAMO directly generates images if flattened images are used, it only generates vectors in the distributed representation space (mapped by the convolutional layers) for the convolutional variant. Therefore, we also propose the GAMO2pix mechanism to obtain images from the GAMO-generated vectors in the distributed representation space. 

\begin{table}[!ht]
    \centering
    \caption{Comparison of FID of cDCGAN and GAMO2pix.}
    \label{tab:fidScores}
    \vskip 0.05in
    \scriptsize
    \begin{tabular}{ccc} \toprule
       Dataset & GAMO2pix (Ours) & cDCGAN \\ \midrule
       Fashion-MNIST & \textbf{0.75$\pm$0.03} & 5.57$\pm$0.03 \\
       SVHN & \textbf{0.17$\pm$0.02} & 0.59$\pm$0.04 \\
       CIFAR10 & \textbf{1.59$\pm$0.03} & 2.96$\pm$0.03 \\
       CelebA-Small & \textbf{11.13$\pm$0.04} & 15.12$\pm$0.05 \\ \bottomrule
    \end{tabular}
\end{table}

Our network for generating images (as illustrated in Figure \ref{fig:gamo2pix}) from the GAMO-generated vectors is inspired by the Variational Autoencoder (VAE) \cite{kingma2013auto, rezende2014auto}. VAE, unlike regular autoencoders, is a generative model which attempts to map the encoder output to a standard normal distribution in the latent space, while the decoder is trained to map samples from the latent normal distribution to images. We follow the design of a standard VAE in GAMO2pix, only replacing the encoder part with the convolutional feature extractor $F$ trained by GAMO. The GAMO2pix network is trained separately for each class while keeping the encoder part fixed. Such a setting should learn the inverse map from the $D$-dimensional feature space induced by $F$ to the original image space and consequently be able to generate realistic images of the concerned class given GAMO-generated vectors for that class as input.


We present the images respectively generated by cDCGAN and GAMO2pix on CIFAR10, Fashion-MNIST, SVHN and CelebA-Small in Figures \ref{fig:cdcganim}-\ref{fig:gamoim}. We can see that GAMO2pix can indeed generate more realistic and diverse images, compared to cDCGAN which also suffers from mode collapse for minority classes. This is further confirmed by the lower Fr{\'e}chet Inception Distance (FID) \cite{martin2017fid} (calculated between real and artificial images from each class and averaged over classes) achieved by GAMO2pix, as shown in Table \ref{tab:fidScores}.

\section{Conclusions and Future Work}\label{sec:concl}
The proposed GAMO is an effective end-to-end oversampling technique for handling class imbalance in deep learning frameworks. Moreover, it is also an important step towards training robust discriminative models using adversarial learning. We have observed from our experiments that the convolutional variant of GAMO is more effective due to the distributed representations learned by the convolutional layers. We also found that the LS loss variant of GAMO generally performs better than the CE loss variant.

An interesting area of future investigation is to improve the quality of the images generated by GAMO2pix by employing a different architecture such as BEGAN \cite{berthelot2017began}. To reduce the tendency of GAMO to overfit as well as to potentially improve its performance, one may consider hybridization with improved GAN variants \cite{gurumurthy2017deliganFuture} which can achieve good performance even with less number of training samples. Further, one may explore the efficacy of GAMO to learn new classes by taking inspiration from Memory Replay GAN \cite{wu2018memoryFuture}, or study the usefulness of the proposed convex generator for handling boundary distortion in GANs.

{\small
\bibliographystyle{ieee}
\bibliography{paper}

\begin{thebibliography}{10}\itemsep=-1pt

\bibitem{shin2017dos}
S.~Ando and C.~Y. Huang.
\newblock Deep over-sampling framework for classifying imbalanced data.
\newblock In {\em Machine Learning and Knowledge Discovery in Databases}, pages
  770--785. Springer International Publishing, 2017.

\bibitem{barua2014mvmote}
S.~{Barua}, M.~M. {Islam}, X.~{Yao}, and K.~{Murase}.
\newblock Mwmote--majority weighted minority oversampling technique for
  imbalanced data set learning.
\newblock {\em IEEE Transactions on Knowledge and Data Engineering},
  26(2):405--425, 2014.

\bibitem{berthelot2017began}
D.~Berthelot, T.~Schumm, and L.~Metz.
\newblock Began: boundary equilibrium generative adversarial networks.
\newblock {\em arXiv preprint arXiv:1703.10717}, 2017.

\bibitem{branco2016}
P.~Branco, L.~Torgo, and R.~P. Ribeiro.
\newblock A survey of predictive modeling on imbalanced domains.
\newblock {\em ACM Computing Surveys (CSUR)}, 49(2):31, 2016.

\bibitem{buda2018systematic}
M.~Buda, A.~Maki, and M.~A. Mazurowski.
\newblock A systematic study of the class imbalance problem in convolutional
  neural networks.
\newblock {\em Neural Networks}, 106:249--259, 2018.

\bibitem{rota2017cost}
S.~R. Bul{\`o}, G.~Neuhold, and P.~Kontschieder.
\newblock Loss max-pooling for semantic image segmentation.
\newblock In {\em Proceedings of the IEEE Conference on Computer Vision and
  Pattern Recognition}, pages 2126--2135, 2017.

\bibitem{chumphol2009sls}
C.~Bunkhumpornpat, K.~Sinapiromsaran, and C.~Lursinsap.
\newblock Safe-level-smote: Safe-level-synthetic minority over-sampling
  technique for handling the class imbalanced problem.
\newblock In {\em Advances in Knowledge Discovery and Data Mining}, pages
  475--482, 2009.

\bibitem{chawla2002smote}
N.~V. Chawla, K.~W. Bowyer, L.~O. Hall, and W.~P. Kegelmeyer.
\newblock Smote: synthetic minority over-sampling technique.
\newblock {\em Journal of artificial intelligence research}, 16:321--357, 2002.

\bibitem{chawla2003smoteboost}
N.~V. Chawla, A.~Lazarevic, L.~O. Hall, and K.~W. Bowyer.
\newblock Smoteboost: Improving prediction of the minority class in boosting.
\newblock In {\em Knowledge Discovery in Databases: PKDD 2003}, pages 107--119,
  2003.

\bibitem{Chung2016c}
Y.-A. Chung, H.-T. Lin, and S.-W. Yang.
\newblock Cost-aware pre-training for multiclass cost-sensitive deep learning.
\newblock In {\em Proceedings of the Twenty-Fifth International Joint
  Conference on Artificial Intelligence}, IJCAI'16, pages 1411--1417. AAAI
  Press, 2016.

\bibitem{das2018handling}
S.~Das, S.~Datta, and B.~B. Chaudhuri.
\newblock Handling data irregularities in classification: Foundations, trends,
  and future challenges.
\newblock {\em Pattern Recognition}, 81:674--693, 2018.

\bibitem{dong2018imbalanced}
Q.~Dong, S.~Gong, and X.~Zhu.
\newblock Imbalanced deep learning by minority class incremental rectification.
\newblock {\em IEEE Transactions on Pattern Analysis and Machine Intelligence},
  2018.

\bibitem{douzas2018effective}
G.~Douzas and F.~Bacao.
\newblock Effective data generation for imbalanced learning using conditional
  generative adversarial networks.
\newblock {\em Expert Systems with applications}, 91:464--471, 2018.

\bibitem{fernandez2018smote}
A.~Fern{\'a}ndez, S.~Garcia, F.~Herrera, and N.~V. Chawla.
\newblock Smote for learning from imbalanced data: Progress and challenges,
  marking the 15-year anniversary.
\newblock {\em Journal of Artificial Intelligence Research}, 61:863--905, 2018.

\bibitem{goodfellow2014generative}
I.~Goodfellow, J.~Pouget-Abadie, M.~Mirza, B.~Xu, D.~Warde-Farley, S.~Ozair,
  A.~Courville, and Y.~Bengio.
\newblock Generative adversarial nets.
\newblock In {\em Advances in neural information processing systems}, pages
  2672--2680, 2014.

\bibitem{gurumurthy2017deliganFuture}
S.~Gurumurthy, R.~Kiran~Sarvadevabhatla, and R.~Venkatesh~Babu.
\newblock Deligan: Generative adversarial networks for diverse and limited
  data.
\newblock In {\em Proceedings of the IEEE Conference on Computer Vision and
  Pattern Recognition}, pages 166--174, 2017.

\bibitem{hui2005bSmote}
H.~Han, W.-Y. Wang, and B.-H. Mao.
\newblock Borderline-smote: A new over-sampling method in imbalanced data sets
  learning.
\newblock In {\em Advances in Intelligent Computing}, pages 878--887, 2005.

\bibitem{He2008adasyn}
H.~He, Y.~Bai, E.~A. Garcia, and S.~Li.
\newblock {ADASYN}: Adaptive synthetic sampling approach for imbalanced
  learning.
\newblock In {\em IEEE International Joint Conference on Neural Networks},
  pages 1322--1328, 2008.

\bibitem{He2009learningimb}
H.~He and E.~A. Garcia.
\newblock Learning from imbalanced data.
\newblock {\em IEEE Transactions on Knowledge and Data Engineering},
  21(9):1263--1284, 2009.

\bibitem{martin2017fid}
M.~Heusel, H.~Ramsauer, T.~Unterthiner, B.~Nessler, and S.~Hochreiter.
\newblock Gans trained by a two time-scale update rule converge to a local nash
  equilibrium.
\newblock In {\em Advances in Neural Information Processing Systems 30}, pages
  6626--6637, 2017.

\bibitem{huang2016learning}
C.~Huang, Y.~Li, C.~Change~Loy, and X.~Tang.
\newblock Learning deep representation for imbalanced classification.
\newblock In {\em Proceedings of the IEEE Conference on Computer Vision and
  Pattern Recognition}, pages 5375--5384, 2016.

\bibitem{khan2018cost}
S.~H. Khan, M.~Hayat, M.~Bennamoun, F.~A. Sohel, and R.~Togneri.
\newblock Cost-sensitive learning of deep feature representations from
  imbalanced data.
\newblock {\em IEEE Transactions on Neural Networks and Learning Systems},
  29(8):3573--3587, 2018.

\bibitem{kingma2013auto}
D.~P. Kingma and M.~Welling.
\newblock Auto-encoding variational bayes.
\newblock {\em arXiv preprint arXiv:1312.6114}, 2013.

\bibitem{krawczyk2016learning}
B.~Krawczyk.
\newblock Learning from imbalanced data: open challenges and future directions.
\newblock {\em Progress in Artificial Intelligence}, 5(4):221--232, 2016.

\bibitem{krizhevsky2009cifar}
A.~Krizhevsky.
\newblock Learning multiple layers of features from tiny images.
\newblock Technical Report TR-2009, University of Toronto, 2009.

\bibitem{kubat1997}
M.~Kubat, S.~Matwin, et~al.
\newblock Addressing the curse of imbalanced training sets: one-sided
  selection.
\newblock In {\em Icml}, volume~97, pages 179--186, 1997.

\bibitem{kumar2017semi}
A.~Kumar, P.~Sattigeri, and T.~Fletcher.
\newblock Semi-supervised learning with gans: Manifold invariance with improved
  inference.
\newblock In {\em Advances in Neural Information Processing Systems}, pages
  5534--5544, 2017.

\bibitem{lecun1998mnist}
Y.~LeCun, L.~Bottou, Y.~Bengio, and P.~Haffner.
\newblock Gradient-based learning applied to document recognition.
\newblock {\em Proceedings of the IEEE}, 86(11):2278--2324, 1998.

\bibitem{Lin2013Dys}
M.~Lin, K.~Tang, and X.~Yao.
\newblock Dynamic sampling approach to training neural networks for multiclass
  imbalance classification.
\newblock {\em IEEE Transactions on Neural Networks and Learning Systems},
  24(4):647--660, 2013.

\bibitem{lin2017cost}
T.~Lin, P.~Goyal, R.~Girshick, K.~He, and P.~Dollár.
\newblock Focal loss for dense object detection.
\newblock In {\em 2017 IEEE International Conference on Computer Vision
  (ICCV)}, pages 2999--3007. IEEE, 2017.

\bibitem{mao2017least}
X.~Mao, Q.~Li, H.~Xie, R.~Y. Lau, Z.~Wang, and S.~Paul~Smolley.
\newblock Least squares generative adversarial networks.
\newblock In {\em Proceedings of the IEEE International Conference on Computer
  Vision}, pages 2794--2802, 2017.

\bibitem{mazurowski2008training}
M.~A. Mazurowski, P.~A. Habas, J.~M. Zurada, J.~Y. Lo, J.~A. Baker, and G.~D.
  Tourassi.
\newblock Training neural network classifiers for medical decision making: The
  effects of imbalanced datasets on classification performance.
\newblock {\em Neural networks}, 21(2-3):427--436, 2008.

\bibitem{mirza2014conditional}
M.~Mirza and S.~Osindero.
\newblock Conditional generative adversarial nets.
\newblock {\em arXiv preprint arXiv:1411.1784}, 2014.

\bibitem{netzer2011svhn}
Y.~Netzer, T.~Wang, A.~Coates, A.~Bissacco, B.~Wu, and A.~Y. Ng.
\newblock Reading digits in natural images with unsupervised feature learning.
\newblock In {\em NIPS workshop on deep learning and unsupervised feature
  learning}, page~5, 2011.

\bibitem{odena2017conditional}
A.~Odena, C.~Olah, and J.~Shlens.
\newblock Conditional image synthesis with auxiliary classifier gans.
\newblock In {\em Proceedings of the 34th International Conference on Machine
  Learning-Volume 70}, pages 2642--2651. JMLR. org, 2017.

\bibitem{radford2015unsupervised}
A.~Radford, L.~Metz, and S.~Chintala.
\newblock Unsupervised representation learning with deep convolutional
  generative adversarial networks.
\newblock {\em arXiv preprint arXiv:1511.06434}, 2015.

\bibitem{rezende2014auto}
D.~J. Rezende, S.~Mohamed, and D.~Wierstra.
\newblock Stochastic backpropagation and approximate inference in deep
  generative models.
\newblock {\em arXiv preprint arXiv:1401.4082}, 2014.

\bibitem{salimans2016improved}
T.~Salimans, I.~Goodfellow, W.~Zaremba, V.~Cheung, A.~Radford, and X.~Chen.
\newblock Improved techniques for training gans.
\newblock In {\em Advances in neural information processing systems}, pages
  2234--2242, 2016.

\bibitem{santurkar2018classification}
S.~Santurkar, L.~Schmidt, and A.~Madry.
\newblock A classification-based study of covariate shift in gan distributions.
\newblock In {\em International Conference on Machine Learning}, pages
  4487--4496, 2018.

\bibitem{Sokolova2009}
M.~Sokolova and G.~Lapalme.
\newblock A systematic analysis of performance measures for classification
  tasks.
\newblock {\em Information Processing \& Management}, 45(4):427 -- 437, 2009.

\bibitem{springenberg2015unsupervised}
J.~T. Springenberg.
\newblock Unsupervised and semi-supervised learning with categorical generative
  adversarial networks.
\newblock {\em arXiv preprint arXiv:1511.06390}, 2015.

\bibitem{srivastava2017veegan}
A.~Srivastava, L.~Valkov, C.~Russell, M.~U. Gutmann, and C.~Sutton.
\newblock Veegan: Reducing mode collapse in gans using implicit variational
  learning.
\newblock In {\em Advances in Neural Information Processing Systems 30}, pages
  3308--3318, 2017.

\bibitem{wang2016d}
S.~Wang, W.~Liu, J.~Wu, L.~Cao, Q.~Meng, and P.~Kennedy.
\newblock Training deep neural networks on imbalanced data sets.
\newblock In {\em 2016 International Joint Conference on Neural Networks
  (IJCNN)}, pages 4368--4374. IEEE, 2016.

\bibitem{wang2017meta}
Y.-X. Wang, D.~Ramanan, and M.~Hebert.
\newblock Learning to model the tail.
\newblock In {\em Advances in Neural Information Processing Systems}, pages
  7029--7039, 2017.

\bibitem{wu2018memoryFuture}
C.~Wu, L.~Herranz, X.~Liu, J.~van~de Weijer, B.~Raducanu, et~al.
\newblock Memory replay gans: Learning to generate new categories without
  forgetting.
\newblock In {\em Advances in Neural Information Processing Systems}, pages
  5966--5976, 2018.

\bibitem{xiao2017fashion}
H.~Xiao, K.~Rasul, and R.~Vollgraf.
\newblock Fashion-mnist: a novel image dataset for benchmarking machine
  learning algorithms.
\newblock {\em arXiv preprint arXiv:1708.07747}, 2017.

\bibitem{xiao2010sun}
J.~Xiao, J.~Hays, K.~A. Ehinger, A.~Oliva, and A.~Torralba.
\newblock Sun database: Large-scale scene recognition from abbey to zoo.
\newblock In {\em Computer vision and pattern recognition (CVPR), 2010 IEEE
  conference on}, pages 3485--3492, 2010.

\bibitem{xie2015holistically}
S.~Xie and Z.~Tu.
\newblock Holistically-nested edge detection.
\newblock In {\em Proceedings of the IEEE international conference on computer
  vision}, pages 1395--1403, 2015.

\bibitem{xie2017holistically}
S.~Xie and Z.~Tu.
\newblock Holistically-nested edge detection.
\newblock {\em International Journal of Computer Vision}, 125(1-3):3--18, 2017.

\bibitem{yan2015}
Y.~Yan, M.~Chen, M.~Shyu, and S.~Chen.
\newblock Deep learning for imbalanced multimedia data classification.
\newblock In {\em 2015 IEEE International Symposium on Multimedia (ISM)}, pages
  483--488. IEEE, 2015.

\bibitem{yu2015lsun}
F.~Yu, A.~Seff, Y.~Zhang, S.~Song, T.~Funkhouser, and J.~Xiao.
\newblock Lsun: Construction of a large-scale image dataset using deep learning
  with humans in the loop.
\newblock {\em arXiv preprint arXiv:1506.03365}, 2015.

\end{thebibliography}
}

\end{document}